\documentclass[manuscript]{acmart}

\setcopyright{none}
\copyrightyear{2023} 
\acmYear{2023} 
\acmConference[FAccTRec 2023]{The 6th Workshop on Responsible Recommendation}{September 18--22, 2023}{Singapore}

\renewcommand\footnotetextcopyrightpermission[1]{} 
\usepackage{booktabs} 

\usepackage{mathsymbols}

\begin{document}

\title{Re-formalization of Individual Fairness}

\author{Toshihiro Kamishima}
\affiliation{%
    \institution{National Institute of Advanced Industrial Science and Technology (AIST)}
    \city{Tsukuba}
    \state{Ibaraki}
    \country{Japan}
}
\email{mail@kamishima.net}

\renewcommand{\shortauthors}{Kamishima}

\maketitle

\section{Introduction}

The notion of individual fairness is a formalization of an ethical principle, ``Treating like cases alike,'' which has been argued such as by Aristotle.
In a fairness-aware machine learning context, Dwork et al.\ firstly formalized the notion.
In their formalization, a similar pair of data in an unfair space should be mapped to similar positions in a fair space.
We propose to re-formalize individual fairness by the statistical independence conditioned by individuals.
This re-formalization has the following merits.
First, our formalization is compatible with that of Dwork et al.
Second, our formalization enables to combine individual fairness with the fairness notion, equalized odds or sufficiency, as well as statistical parity.
Third, though their formalization implicitly assumes a pre-process approach for making fair prediction, our formalization is applicable to an in-process or post-process approach.

\section{Fairness in Machine Leanring}



We here introduce a theory of fairness surveyed by Lippert-Ras\-mu\-ssen~\cite{ej:071}, and relates the theory to fairness in machine learning.
According to the theory, there are two major accounts why a discrimination becomes bad.
In one harm-based account, ``\emph{An instance of discrimination is pro tanto bad, when it is, because it makes the discriminatees worse off}.''
In another disrespect-based account, ``\emph{An act or practice is morally disrespectful of X if, and only if, it in some way presupposes that X has a lower moral status than he or she in fact has}.''
While harm-based and disrespect-based accounts connected to the utilitarianism and the Kantianism, respectively.
These accounts are different, for example, in a point that, ``\emph{While the harm-based account of the badness of discrimination implies a moral symmetry between discriminating against someone and discriminating in favor of someone, no such symmetry exists on the disrespect-based account}.''

We consider that a fairness-aware machine learning tries to resolve unfairness explained by the harm-based account, due to the following reasons.
First, representing people by quantitative measures is occasionally claimed as a disrespectful act.
In this case, machine learning techniques cannot be applied.
Second, fairness-aware techniques are developed to meet legal fairness concepts related to the harm-based account.
For example, \emph{a gross statistical parity}~[Hazelwood School District v. United States, 433 U.S. 299 (1977)]~\cite{xj:0573} or \emph{a but-for cause}~[Jack Gross, Petitioner, v. FBL Financial Services, US Supreme Court, 2008]~\cite{misc:291}.

Lippert-Rassmussen further discussed a baseline for determining whether a discriminatee has been made worse off~\cite{ej:071}.
One baseline is a counterfactual that a discriminatee had not been discriminated.
Another is an ideal outcome that a discriminatee would have been in a just outcome.
In a fairness-aware ML context, we consider that the former case corresponds to counterfactual fairness~\cite{neurips:17:03}, and that association-based fairness uses the later baseline.
In this paper, we focus on association-based fairness.

\subsection{Association-Based Fairness}

We show our notations and what is formal fairness.
$Y$ denotes a target variable, which represents an objective of decision making, such as loan approval or university admission.
While $Y$ is an observed outcome, $\hat{Y}$ is an estimated outcome.
$S$ denotes a sensitive feature, which represents socially sensitive information, such as a gender or race.
$\bfX$ is a non-sensitive feature vector, which corresponds to all features other than a sensitive feature.
Formal fairness is defined as an ideal formal relation between target variable, sensitive feature, and non-sensitive features.
As the formal relation, associations between variables is used in association-based fairness.
Correlation has been used as association in early studies until 1970s, but fairness is recently discussed based on independence~\cite{facct:19:01}.

We distinguish fairness criteria from fairness measures.
The criteria is a condition defined by using independence, and the measures quantify how exactly the criteria are satisfied.
For example, a widely-used criterion, statistical parity~\cite{ec:044}, is defined by the independence between an estimated target variable and a sensitive feature,
\begin{equation}
\hat{Y} \indep S
.
\label{eq:sp}
\end{equation}
How exactly this criterion is satisfied is measured, for example, by a $\chi^2$ statistic~\cite{macm:13:01}, mutual information~\cite{epublist:127}, or balanced error ratio~\cite{kdd:15:01}.
Though so many fairness measures has been proposed~\cite{dmkd:17:01}, we can simplify our discussion in this paper by focusing on criteria.

\section{Individual Fairness}

After showing formalization of individual fairness by Dwork et al., we propose our new formalization, and show these two formalizations are compatible.

Individual fairness is one of fairness criteria, and Dwork et al.\ formalized as follows~\cite{ec:044}.
In their framework, they introduced a map, $M(\cdot): \Dom(\bfX) \to \Dom(\bfX^\circ)$, where $\Dom(\bfX)$ and $\Dom(\bfX^\circ)$ are an original feature space, which is potentially unfair, and a fair feature space, respectively.
Data represented in the fair space allows to make fair prediction.
Then, to translate the principle, ``Treating like cases alike,'' they restrict the map a similar pair of data in an original space must similar also in a fair space.
This is formally a Lipschitz condition:
\begin{equation}
d^\circ(M(\bfx_1), M(\bfx_2)) \le d(\bfx_1, \bfx_2), \forall \bfx_1, \bfx_2 \in \Dom(\bfX)
,
\label{eq:Lipschitz}
\end{equation}
where $\bfx_i$ is an instance of $\bfX$, and $d(\cdot,\cdot)$ and $d^\circ(\cdot,\cdot)$ are distances in original space and fair space, respectively.
Hence, like cases in an original space are treated alike in a fair space.

We propose to re-formalize the principle.
We first interpret that the principle implicitly assumes the condition regardless a sensitive feature.
If not, unlike cases regarding a sensitive feature might be treated unlikely.
This is formalized as
\begin{equation}
\Pr[\hat{Y} \mid S, \bfx] = \Pr[\hat{Y} \mid \bfx],\;\forall\bfx
,
\label{eq:if1}
\end{equation}
where $\bfx$ is an instance of a random variable $\bfX$.
An instance, $\bfx$, is assumed to contain the information relevant to an individual except for the individual's sensitive information, and $\bfx$ is considered as a representation of the individual.
Because the Eq.\eqref{eq:if1} guarantees that its outcome, $\hat{Y}$, follows the same distribution for the same $\bfx$, we claim that Eq.\eqref{eq:if1} represents the principle.
Furthermore, by definition, Eq.\eqref{eq:if1} is equivalent to the conditional independence:
\begin{equation}
\hat{Y} \indep S \mid \bfX
.
\label{eq:if2}
\end{equation}

We finally show that our formalization of individual fairness is compatible with that of Dwork et al.
For this proof, we show that both formalization is compatible with another formal fairness condition, fairness through unawareness~\cite{e:0085,ec:044}.
This fairness through unawareness is to simply omit a sensitive feature in prediction model, and a prediction model satisfies:
\begin{equation}
\Pr[\hat{Y} \mid S, \bfX] = \Pr[\hat{Y} \mid \bfX]
\label{eq:ftu1}
.
\end{equation}
This is, by definition, equivalent to the conditional independence:
\begin{equation}
\hat{Y} \indep S \mid \bfX
.
\label{eq:ftu}
\end{equation}
This is formally coincident with Eq.\eqref{eq:if2}, and our formalization of individual fairness is compatible with fairness through unawareness.
We move on to a case of Dwork et al.
In their distance measure, no sensitive information is taken into account, and the information is eliminated in a fair space.
Hence, they implicitly omit the sensitive information in prediction, and their method satisfies a condition of fairness through unawareness.
From the above discussion, it is reasonable to conclude that our formalization of individual fairness is compatible with that of Dwork et al.

It is worth to note a legal notion of \emph{situation testing} that is formalized by Luong~et~ al.~\cite{kdd:11:01}.
They employed legally-grounded features, which are allowed to use for decisions by law.
They are similar to non-sensitive features, but they are assumed that all the features are assumed to be legal if they are not sensitive.
If the assumption is adopted, the situation testing can be treated as a kind of individual fairness, and it can be formalized as conditioning by $\bfX$ as above.

\subsection{Extension of Individual Fairness}

We first discuss another interpretation of our formalization.
Comparing statistical parity, Eq.\eqref{eq:sp}, with individual fairness, Eq.\eqref{eq:if2}, Eq.\eqref{eq:if2} can be derived by conditioning Eq.\eqref{eq:sp} by $\bfX$.
As described above, each instance of $\bfX$ represents an individual.
Hence, we can interpret that Eq.\eqref{eq:sp} satisfies statistical parity for each individual.
In particular, statistical parity is satisfied for each individual, having the same attributes or abilities.
Hence, it would be appropriate to treat Eq.\eqref{eq:if2} as \emph{individual statistical parity}.

If we extend a notion of individual fairness by treating it as conditioning a fairness criterion by $\bfX$, we can create an individualized version of the criterion.
Here, we discuss two widely-used fairness criteria.
One is equalized odds~\cite{neurips:16:02}:
\begin{equation}
\hat{Y} \indep S \mid Y
.
\label{eq:eo}
\end{equation}
In a case that all variables are binary, this is equivalent to match false positive rate and false negative rates simultaneously.
Another is sufficiency~\cite{e:0085} or calibration~\cite{ec:051,ej:078}:
\begin{equation}
Y \indep S \mid \hat{Y}
.
\label{eq:suff}
\end{equation}
In a case that all variables are binary, this is equivalent to match positive and negative predictive values simultaneously.
These fairness conditions are widely known that the ProPublica pointed out the recidivism score, the COMPAS, does not satisfy equalized odds~\cite{misc:219}.
However, the US Court and a developer refuted that the score is designed to satisfy a sufficiency condition~\cite{ej:073,misc:318}.

Conditioning by $\bfX$ enables to convert these two fairness criteria to individual versions of them.
In this case, the phrase, ``treating alike,`` means predicting in similar error rate.
Then, \emph{individual equalized odds} is
\begin{equation}
\hat{Y} \indep S \mid Y, \bfX
\label{eq:ieo}
,
\end{equation}
and \emph{individual sufficiency} is
\begin{equation}
Y \indep S \mid \hat{Y}, \bfX
.
\label{eq:isuff}
\end{equation}
These two individual criteria would be useful when the parity of errors is needed at an individual level.

As described above, criteria designed so that fairness is maintained at a group level can be convert to the corresponding criteria at an individual level by conditioning non-sensitive features.

\subsection{Application to Other Types of Approaches}


There are three types of approaches to make fair prediction~\cite{jacm:10:01}.
In a pre-process approach, data are transformed into fair representations, and standard predictors are applied to the representations, e.g.~\cite{kdd:15:01}.
Methods of an in-process approach directly make fair prediction, e.g.~\cite{epublist:127}.
After learning a standard predictor, its outcomes are modified so as to be fair in a post-process approach, e.g.~\cite{neurips:16:02}.

An approach of Dwork et al.\ is a kind of a pre-process, and their formalization of individual fairness premises the approach.
However, re-formalizing individual fairness permits to use criteria of individual fairness in in-process or post-process approaches.
This would make individual fairness applicable in a more broad way.

\section{Conclusion}

\begin{table*}
\centering
\caption{Summary of Fairness Criteria}
\label{tab:criteria}
\begin{tabular}{%
p{0.10\linewidth}@{\makebox[0.03\linewidth]{}}%
p{0.13\linewidth}@{\makebox[0.01\linewidth]{}}%
p{0.13\linewidth}@{\makebox[0.01\linewidth]{}}%
p{0.13\linewidth}@{\makebox[0.03\linewidth]{}}%
p{0.13\linewidth}@{\makebox[0.01\linewidth]{}}%
p{0.13\linewidth}@{\makebox[0.01\linewidth]{}}%
p{0.13\linewidth}@{\makebox[0.01\linewidth]{}}%
}\toprule
&
statistical parity & 
equalized odds &
sufficiency &
individual statistical parity & 
individual equalized odds &
individual sufficiency \\\midrule
condition &
$\hat{Y} \indep S$ &
$\hat{Y} \indep S \mid Y$ &
$Y \indep S \mid \hat{Y}$ &
$\hat{Y} \indep S \mid \bfX$ &
$\hat{Y} \indep S \mid Y, \bfX$ &
$Y \indep S \mid \hat{Y}, \bfX$ \\
unit & group & group & group & individual & individual & individual \\
awareness & aware & aware & aware & unaware & aware & aware \\\bottomrule
\end{tabular}%
\end{table*}

We re-formalize the notion of individual fairness.
Our formalization is compatible with that of Dwork et al., and this formalization enables to create individual version of equalized odds or sufficiency.
Our new individual fairness can be used also in in-process or post-process approaches.
Finally, we summarize fairness criteria discussed in this paper in Table~\ref{tab:criteria}.

One of the limitation of the above discussion is an interpretation of the term, \emph{like}.
In this paper, if non-sensitive features take exactly the same values, two assumptive individuals are considered as \emph{like}.
If non-sensitive features are nominal and discrete, this interpretation would not be so problematic.
However, if they are continuous, a small difference in non-sensitive features between two individuals might be ignored and these individuals should be considered as \emph{like}.
In this case, a soft version of conditioning might be useful.
For example, a condition of individual statistical parity is satisfied in the neighbor of a specific individual, $\bfx$, with high probability. 


\begin{acks}
This work is supported by MEXT/JSPS KAKENHI Grant Number JP24500194, JP15K00327, 18H03300, and 21H03504.
\end{acks}

\bibliographystyle{ACM-Reference-Format}
\bibliography{main}

%
%

\end{document}